\documentclass[lettersize, journal]{IEEEtran}
\usepackage{amsmath,amsfonts}
\usepackage{algorithmic}
\usepackage{algorithm}
\usepackage{array}
\usepackage[caption=false,font=normalsize,labelfont=sf,textfont=sf]{subfig}
\usepackage{textcomp}
\usepackage{stfloats}
\usepackage{url}
\usepackage{verbatim}
\usepackage{graphicx}
\usepackage{cite}
\usepackage{float}
\usepackage{multirow}
\usepackage[margin=1.78cm]{geometry}
\usepackage[switch]{lineno}
\usepackage{multicol}
\usepackage{bm}
\usepackage{makecell} 
\usepackage{amsmath}
\DeclareMathOperator*{\minimize}{Minimize}

\usepackage{optidef}
\usepackage{mathrsfs} 

\begin{document}

\title{Data-Efficient Modeling for Precise Power Consumption Estimation of Quadrotor Operations Using Ensemble Learning}

\author{Wei Dai, Mingcheng Zhang, and Kin Huat Low
\thanks{This research is supported by the National Research Foundation, Singapore, and the Civil Aviation Authority of Singapore, under the Aviation Transformation Programme. The second author’s candidature
scholarship is provided by the Nanyang Technological University through Air Traffic Management Research Institute Leadership Track. The authors would like to thank Mr. Bizhao Pang for his assistance in the flight tests. Any opinions, findings and conclusions or recommendations expressed in this material are those of the authors and do not reflect the views of National Research Foundation, Singapore and the Civil Aviation Authority of Singapore.}
\thanks{Wei Dai (e-mail: wei.dai@ntu.edu.sg) and Mingcheng Zhang (e-mail: m200057@e.ntu.edu.sg) are with the School of Mechanical and Aerospace Engineering, Nanyang Technological University, Singapore 639798, and also with the Air Traffic Management Research Institute, Nanyang Technological University, Singapore 637460.}
\thanks{Kin Huat Low (corresponding author, e-mail: mkhlow@ntu.edu.sg) is with the School of Mechanical and Aerospace Engineering, Nanyang Technological University, Singapore 639798.}}

\markboth{Journal of \LaTeX\ Class Files,~Vol., No., May~2022}%
{Shell \MakeLowercase{\textit{et al.}}: A Sample Article Using IEEEtran.cls for IEEE Journals}


\maketitle

\begin{abstract}
Electric Take-Off and Landing (eVTOL) aircraft is considered as the major aircraft type in the emerging urban air mobility. Accurate power consumption estimation is crucial to eVTOL, supporting advanced power management strategies and improving the efficiency and safety performance of flight operations. In this study, a framework for power consumption modeling of eVTOL aircraft was established. We employed an ensemble learning method, namely stacking, to develop a data-driven model using flight records of three different types of quadrotors. Random forest and extreme gradient boosting, showing advantages in prediction, were chosen as base-models, and a linear regression model was used as the meta-model. The established stacking model can accurately estimate the power of a quadrotor. Error analysis shows that about 80\% prediction errors fall within one standard deviation interval and less than 0.5\% error in the prediction for an entire flight can be expected with a confidence of more than 80\%. Our model outperforms the existing models in two aspects: firstly, our model has a better prediction performance, and secondly, our model is more data-efficient, requiring a much smaller dataset. Our model provides a powerful tool for operators of eVTOL aircraft in mission management and contributes to promoting safe and energy-efficient urban air traffic.

\end{abstract}


\section{Introduction}
\IEEEPARstart{I}{n} recent years, traffic congestion has become a major concern in megacities. Stakeholders started to look for solutions to extend the urban traffic system to the third dimension. The demand for urban air mobility (UAM) is emerging, which attracts the attention of authorities, and industrial companies. An increasing number of parties have started investing in UAM~\cite{Thipphavong:2018, Sun2021a}. UAM is bringing new airspace users into the airspace system, these new comers are bringing challenges to the current air traffic system~\cite{LIU2021}. To support safe and efficient UAM operation, researchers have been working on the related fields, including concept of operations for unmanned aircraft system traffic management (UTM)~\cite{Prevot2016}, urban airspace accessibility~\cite{9256427} and management strategy~\cite{Pang2020b}, path/trajectory planning and scheduling algorithms development~\cite{Dai2021, Wu2021}, conflict detection and resolution~\cite{Guan2020}, etc. 

EVTOL aircraft uses electrical power to hover, take-off and land vertically. Due to its high kinematic flexibility and small required region for take-off and landing, the eVTOL aircraft are suitable for operations in complex environments compared with traditional fixed-wing aircraft~\cite{Shao2021}. Typical eVTOL aircraft includes both passenger-carrying aircraft that are under development by many companies like EHang and Volocopter, and drones, which have been very popular in building inspection, parcel delivery, and recreational photography. 

The features of eVTOL aircraft lead to great potential in its application in UAM. Being electrically powered has many advantages, including low-cost and environmentally friendly. Nevertheless, using electricity as the main power source makes energy management a key issue in the development and application of eVTOL aircraft~\cite{Townsend2020}. Due to the limit of battery capacity, the maximum cruising range of eVTOL aircraft is normally smaller than that of a traditional aircraft with fossil fuel engines. Hence the maximum cruising range becomes a crucial constraint in eVTOL flight planning. 

In the absence of commercialized passenger-carrying eVTOL, in this section, we take drones as the subject of research due to the maturity of the aircraft system, the accessibility of aircraft specifications, and simplicity in carrying out experiments. A typical operational scenario of a drone operation in the urban airspace is parcel delivery. The logistic company needs to design a flight path for every delivery mission. Manufacturers of most of the off-the-shelf commercial drones provide the longest flight time as a reference. However, there are many factors affecting the instantaneous power consumption when the drone is airborne, which makes the reference value less reliable. Operators plan flights based on the claimed maximum flight time. Once the actual power consumption is higher than the estimated, there will be a risk of crashing posed to the flight mission. Therefore, an accurate estimation of the power of aircraft energy consumption will play a significant role in promoting efficiency, and more importantly, safety, in flight operations.

\subsection{Literature Review}
A number of existing studies build the power consumption model (PCM) on a statistical basis. These estimations were considered by many researchers in the design and optimization problems for drone deliveries. These topics majorly exist in operations researches, including deposit and/or recharging facility locationing problems~\cite{Hong2018, Shavarani2018, Chauhan2019}, scheduling and path/trajectory planning problems~\cite{Dorling2017, Agatz2018, Schermer2019, Liu2019cor, Cheng2020a}, and truck-drone combined delivery operations~\cite{Chung2020}, etc. According to Zhang et al.~\cite{Zhang2021}, the PCMs built in these studies can be taxonomized as integrated models and component models. In the integrated models, a single parameter, the lift-to-drag ratio, was used to summarize the structural design and the aerodynamics of an aircraft. D’Andrea et al.~\cite{DAndrea2014} carried out the seminal integrated model, and some following contributors of integrated models include Figliozzi~\cite{Figliozzi2017}, Troudi et al. \cite{Troudi2018}, and Gulden~\cite{Gulden2017}. The other group of PCMs is using component models. An early study of this approach was carried out by Dorling et al.~\cite{Dorling2017}, in which the authors performed an equation of the power consumed to maintain a hovering state. Stolaroff et al.~\cite{Stolaroff2018} built a two-component model including the thrust to balance weight and the parasitic drag. Liu et al.~\cite{Liu2017} built a three-component model including the power to conquer weight, parasitic drag, and the drag generated on propellers in rotation. These component models were all developed assuming a steady state of the aircraft (hovering or moving in a fixed direction with a constant speed). Zhang et al.~\cite{Zhang2021} compared and assessed the aforementioned methods, revealed the great differences among the results of different models, and emphasized the significance of calibrating model parameters in flights. The studies mentioned above are modeling the energy consumption towards estimating the energy needed for an entire delivery mission with statistical method. A consideration of the instantaneous variances in power consumption is not included. In the complex urban environment, frequent changes in the direction and magnitude of velocity can be expected because of either urban wind field interruption or confliction avoidance. The change in the kinematics state of aircraft will the power of aircraft energy consumption deviate from steady-state and reduce the accuracy of these models. 

A potentially better way of solving this is to model instantaneous power consumption given the motion of the aircraft is using data-driven machine-learning method. There are a few attempts in this application. Prasetia et al.~\cite{Prasetia2019} used the Elastic Net Regression model to predict the power consumption of a self-built hexacopter drone. The model partially utilizes the kinematic features in each flight: the duration of acceleration and deceleration in each flight. But the eventual output is the power consumption for a flight mission, and the model cannot estimate the momentary power. Tseng et al.~\cite{Tseng2017, Alyassi2017} used a nine-term non-linear regression model to estimate power consumption based on flight data from field experiments. The models built in these two works are too simple to capture the dyanmics of the aircraft. Hence the accuracy of their prediction is limited. Two recent studies used deep learning to build the model. Hong et al.~\cite{Hong2020} used a multi-layer perceptron to model the power consumption of a DJI M600 aircraft. The model was trained by a dataset containing the record of 30 flight hours. Choudhry et al.~\cite{Choudhry2021} used machine learning to model the instantaneous power consumption rate of a quadrotor drone in a series of flight missions. They developed a PCM based on Temporal Convolutional Network (TCN), which was trained by their open-source drone delivery dataset~\cite{Rodrigues2021}. The weakness of these two deep learning model is their dependency on the datasize. In both studies, the dataset contains more than 20 hours of flight record. The dependency on large dataset limits the model's scalability. To the best of our knowledge, there are only a few developments of PCM in the literature, the estimation errors in the results were not extensively discussed. 

In summary, a method that can accurately model the mechanism of how aircraft motions are linked to instantaneous power consumption by using a data-driven machine learning method is absent. 

Ensemble learning is a class of machine learning methods that ensembles multiple base models to obtain a better performance~\cite{Zhang2012}. The ensemble learning algorithms generally fall into three categories: bagging, boosting, and stacking. Different from bagging and boosting models, the stacking models are usually constructed by heterogeneous base-models. This feature allows stacking models to integrate different advantages from heterogeneous models and increases the performance of modeling~\cite{Smyth1999}. In recent years, ensemble learning is widely applied in similar problems, including software defect prediction~\cite{matloob2021software}, wind energy prediction~\cite{da2021novel}, capacity estimation of lithium-ion batteries~\cite{shen2020deep}, data missing estimation~\cite{zhang2021new}, etc. Applying ensemble learning in the estimation of power consumption can potentially lead to a promising performance.



\subsection{Contributions}

The main outcome of this study is the establishment of a data-driven model for estimating the power consumption of a quadrotor in flight operations. The contribution of this study is threefold:
\begin{enumerate}
    \item Our model outperforms the existing models in all three major performance metrics.
    \item Compared with existing methods, our model requires a much smaller dataset, in specific, only 3,000 instances are required in training the model to obtain a good prediction performance, which significantly reduces the efforts needed in the flight tests for data acquisition.
    \item To guarantee the scalability of this study, the input features of the model include only the aircraft’s mass, kinematic state of aircraft, and wind speed. This setting frees the model from enormous and complex details of the aircraft specifications. The developed model can easily be applied to the study of other aircraft types. 
\end{enumerate}

\subsection{Organization}
The paper is organized as follows: In Section~\ref{Sec.Meth}, an overview of the methodology of this study is presented. In Section~\ref{Sec.Data}, the process of data acquisition and cleaning are introduced. In Section~\ref{Sec.Model}, details in the development of the ensemble learning model are explained. In Section~\ref{Sec.Result}, the modeling results are analyzed and compared with the existing models in the literature. And Section~\ref{Sec.Con} concludes the findings in this paper.

\section{Methodology}
\label{Sec.Meth}
The goal of this study is to develop a data-driven PCM as a tool for flight operators. Therefore features as input to the PCM will be major elements of a flight plan, namely the aircraft kinematic states and key environmental factors. The selected features include the aircraft's mass defined as $mass$, the aircraft velocity in the earth-fixed frame defined as $\bm{v} = [v_n, v_e, v_d]$, the linear acceleration in the earth-fixed frame defined as $\dot{\bm{v}} = [\dot{v}_n, \dot{v}_e, \dot{v}_d]$, the Euler angles in earth-fixed frame defined as $\bm{\Theta} = [r, p, y]$, the angular rates vector in the earth-fixed frame defined as $\dot{\bm{\Theta}}_i = [\dot{r}, \dot{q}, \dot{y}]$, and the horizontal wind velocity vector defined as $\bm{w} = [w_e, w_n]$. Thus the input vector to the model is $\bm{x}_i = [mass_i, \bm{v}_i, \dot{\bm{v}}_i, \bm{\Theta}_i, \dot{\bm{\Theta}}_i, \bm{w}_i]$, and $\bm{x}_i \in \mathbb{R}^{1 \times 15}$. And the label $y_i \in \mathbb{R}$ is the power of aircraft, i.e. $y_i = P_i$. By using the training dataset with $m$ input vectors, the total input is $\bm{X} \in \mathbb{R}^{m \times 15}$, and the total ground truth vector is $\bm{y} \in \mathbb{R}^{m \ times 1}$.

The PCM development problem is to select and train a data-driven model $F(\cdot)$ which minimizes the total loss $L$:
\begin{equation}
    \displaystyle{\minimize_{F,\theta}} \quad L(\bm{X}, \bm{y}, F, \bm{\theta})
\end{equation}
where $\theta$ is the parameters of model $F(\cdot)$ that needs to be optimized. The input of the model is the feature vector $\bm{x}_i$, and the output is a prediction $\hat{y}_i$, so that: 
\begin{equation}
    \hat{y}_i = F(\bm{x_i}, \theta)
\end{equation}
We select mean squared error (MSE) as the model's prediction loss. To overcome the overfitting problem that occurs in many machine-learning models, L1 and L2 regularizations are also taken into consideration. Thus the total loss function is defined as:
\begin{equation}
	\begin{split}
	L(\bm{X}, \bm{y}, F, \bm{\theta}) &=  L_{MSE}(\bm{X}, \bm{y}, F, \bm{\theta}) + L1 + L2\\
	&= \frac{1}{m}\sum_i^m (y_i - \hat{y}_i)^2 + \alpha ||\theta|| + \frac{1}{2} \lambda ||\theta||^2
	\end{split}
\end{equation}

\begin{figure}[htb]
    \centering\includegraphics[width=\linewidth]{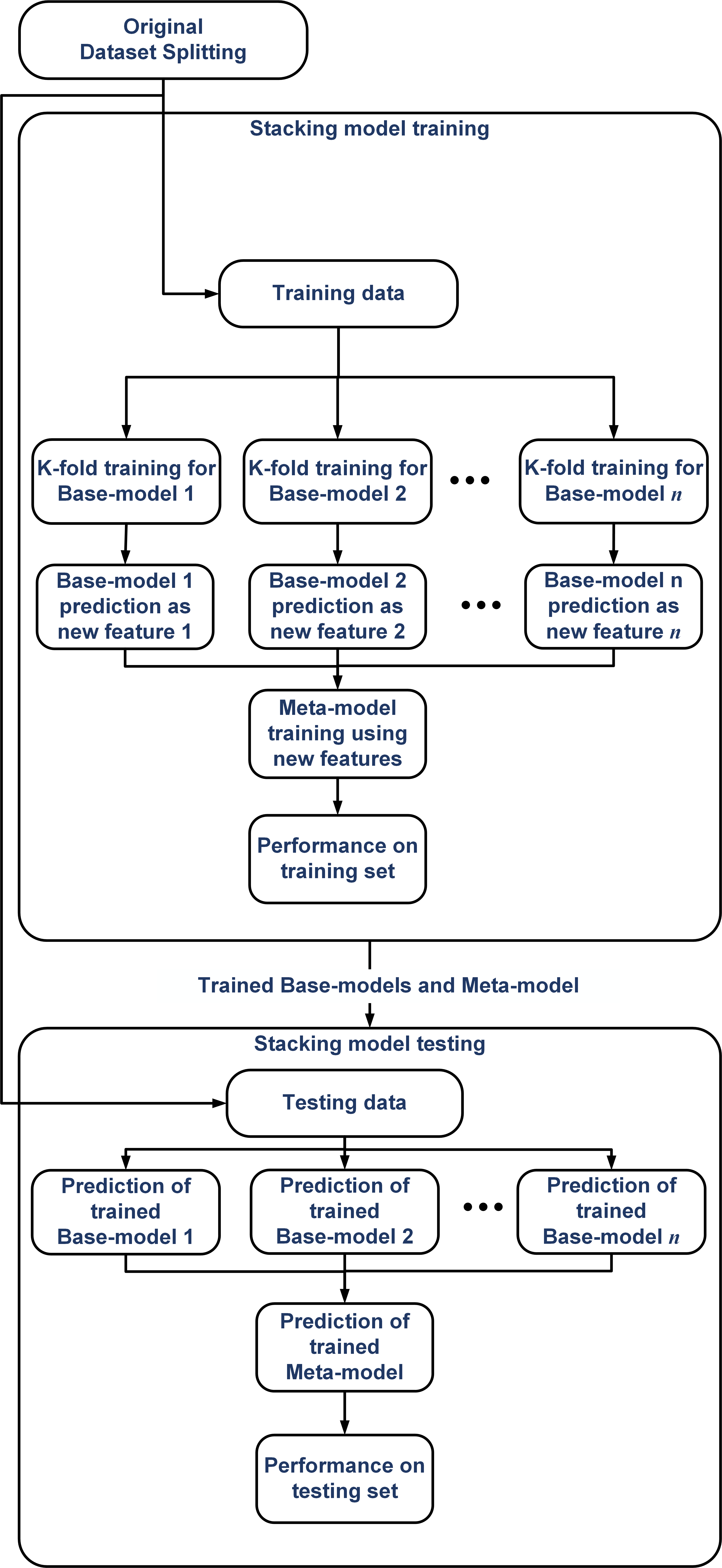}
    \caption{Principle of the training and validation of stacking model}
    \label{stack}
\end{figure}

In this study, we will use an ensemble learning method to model the power of energy consumption by a quadrotor. The major framework of the method is a stacking model. The principle of the stacking model is illustrated in Fig.~\ref{stack}. It consists of two layers; the first layer contains several base-models, and the second layer is a meta-model. Theoretically more layers can be stacked into the model to improve the model’s performance. However, normally the benefit of more than two layers is limited while the additional layer significantly increases the likelihood of overfitting. Hence the basic two-layer structure is selected. In the training phase, K-fold training is used by each base-model. The training set was equally separated to K subsets (K equals to five in the figure). At each time, the model is trained with K-1 subsets and prediction for the rest subset was made. By repeating this procedure for K times, prediction for the entire training set can be achieved. Then the predictions the base-models were used as the new features to train the meta-model. Therefore the entire stacking model is trained and the performance on training set can be gain. Then the trained stacking model is applied to the testing dataset to estimate the generalization performance of the model.

\begin{figure}[t]
    \centering\includegraphics[width=\linewidth]{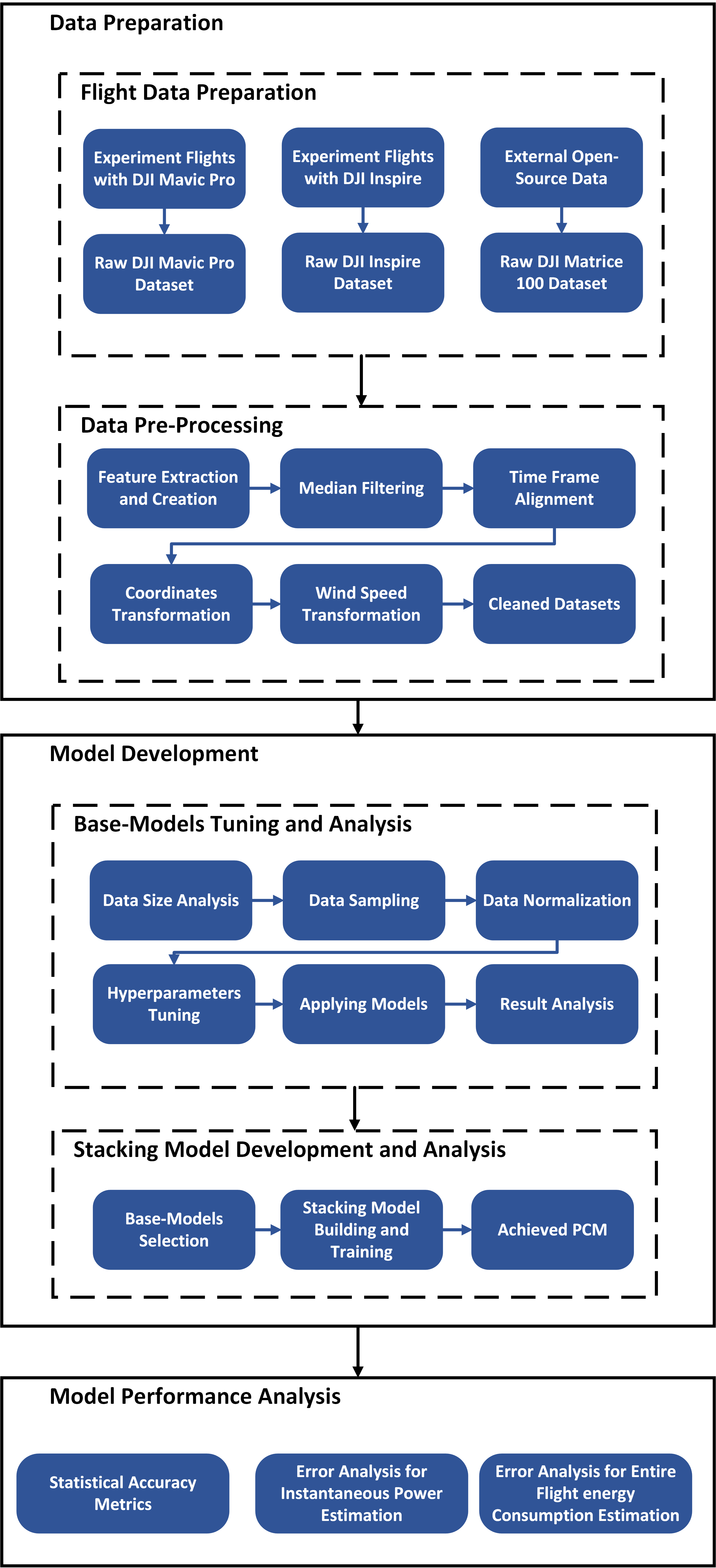}
    \caption{Overall workflow of modeling the power of energy consumption rate of a quadrotor}
    \label{chap03_images:methodology}
\end{figure}

The overall methodology of this study is illustrated in Fig.~\ref{chap03_images:methodology}. To support this data-driven method, three datasets referring to three different aircraft types were prepared. Before modeling, the data was pre-processed to obtain well-cleaned inputs for the models. The stacking model was established based on detailed tuning and characterization of base-models to promote the modeling accuracy. Finally, the performance of the established model was analyzed. 

\section{Data Preparation and Pre-Processing}
\label{Sec.Data}
\subsection{Raw Data Preparation}
There are two sources of the raw dataset used in this study. The first part of the dataset was collected by our research team through a series of flight tests. The aircraft employed in this study were DJI Mavic Pro and DJI Inspire. The accumulated flight times for Mavic Pro and Inspire were 4,994 seconds and 4,317 seconds, respectively. 

The second part of the data was from an open-source dataset published by Rodrigues et al [32]. The authors used a DJI Matrice 100 to perform delivery flights. Different from Mavic Pro and Inspire, Matrice 100 is designed for developers and can carry payloads to capture flight data with different take-off weights. The dataset contains the records of 279 flights. The specifications of Matrice 100 in comparison with Mavic Pro and Inspire are shown in Table~\ref{drone_specs}. 

\begin{table}[H]
    \begin{center}
    \caption{Specifications of the aircraft types concerned in this study.}
    \label{drone_specs}
    \begin{tabular}{|l|c|c|c|}
    \hline
    \textbf{}                  & \multicolumn{1}{|l|}{\makecell[c]{\textbf{DJI}\\\textbf{Mavic Pro}}} & \multicolumn{1}{l}{\makecell[c]{\textbf{DJI}\\\textbf{Inspire}}} & \multicolumn{1}{|l|}{\makecell[c]{\textbf{DJI}\\\textbf{Matrice 100}}} \\ \hline
    Weight (g)                 & 734                                        & 2845                                     & 3680                                         \\
    \hline
    Payload (g)                & -                                          & -                                        & \{0, 250, 500,   750\}                       \\
    \hline
    Max speed (m/s)            & 18                                         & 22                                       & 22                                           \\
    \hline
    Max ascent rate (m/s)  & 5                                          & 5                                        & 5                                            \\
    \hline
    Max descent rate (m/s) & 3                                          & 4                                        & 4                                            \\
    \hline
    Max flight time (min)      & 27                                         & 18                                       & 22                                           \\ \hline
    \end{tabular}
\end{center}
\end{table}

\subsection{Data Pre-Processing}
Careful data pre-processing was carried out before modeling. In the first place, feature extraction was performed. This study aims at estimating the power of energy consumption for the purpose of flight trajectory planning. Hence the kinematic state of aircraft and environmental factors were necessary, which includes the aircraft’s velocity and acceleration, aircraft attitude, angular velocity, and the speed and direction of the wind. Some of the features are not included in the original flight log and thus need to be created. For instance, the GPS output of Mavic Pro and Inspire only has timestamp and location. Thus time-differential method was applied to compute the velocity and acceleration of aircraft. 

Median filtering was used to remove high-frequency noises of the sensors. Onboard sensors’ heterogeneity results in the differences in the frequency of their outputs, as summarized in Table~\ref{sensor_freq}. It is worth noting that in the open-source data for Matrice 100, the position, velocity, and acceleration data are not raw sensor outputs, but were processed data using GPS and IMU outputs and Kalman Filter. Thus the frequency of the aircraft kinematic states in this dataset was 10 Hz. For the unity of data structure, time frame alignment was carried out to merge the step length of all data to 1 second. 

\begin{figure}[H]
      \centering
    	\subfloat[Oiginal dataset] {\includegraphics[width=0.22\textwidth]{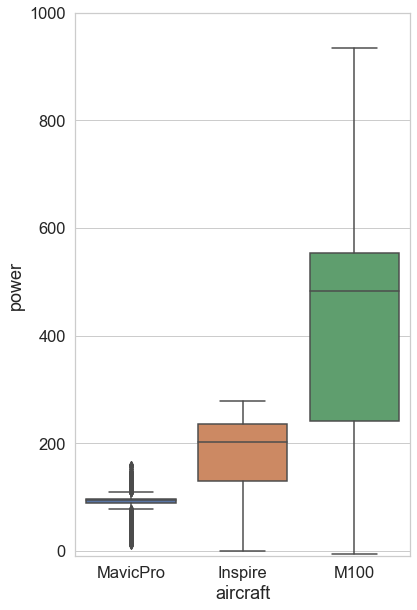}} \quad
        \subfloat[Filtered dataset] {\includegraphics[width=0.22\textwidth]{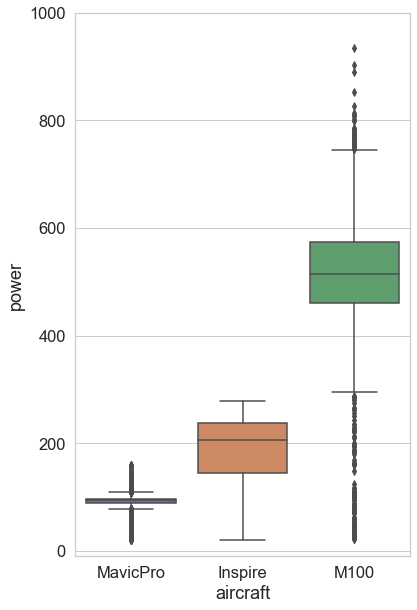}} 
      \caption{Box-plot of the power measurements from the dataset of three aircraft types}
    	\label{power_box}
    	\vspace{0.2in}
\end{figure}

\begin{figure}[H]
    \centering\includegraphics[width=0.8\linewidth]{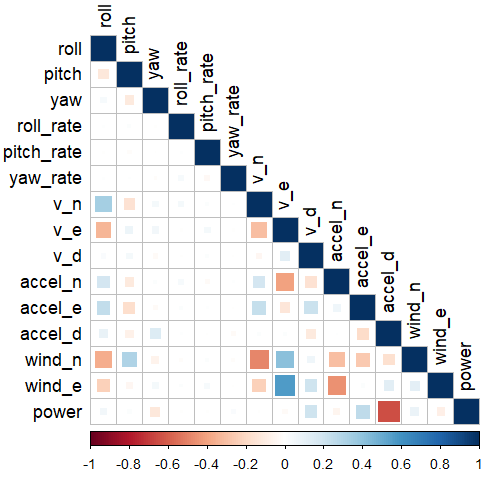}
    \caption{Heat map of pair-wise correlations of the variables in the dataset}
    \label{correlation}
\end{figure}

After transforming all features to the environmental coordinate system, the datasets were well cleaned for modeling. A glance at the data is provided in Fig.~\ref{power_box}~and~\ref{correlation}. The power of the three aircraft used in this study spans from 0 up to 900 Watts during the flight tests, where the heavier the aircraft, the more power consumed during the flights, as illustrated in Fig.~\ref{power_box}. Fig.~\ref{power_box}(a) shows the distribution of power for the three aircraft types in the original dataset. The data was automatically collected when the aircraft is powered on, including the moments that the aircraft is on the ground, when the power consumption is close to zero. By filtering out the power values under 20 Watts, which is insufficient to provide enough lift for any aircraft to take off, the distributions of power are illustrated in Fig.~\ref{power_box}(b). Approximately symmetrical distributions on both sides of the median are observed for all aircraft types. 

\begin{table}[H]
\begin{center}
\caption{Frequency of the output of different sensors.}
\label{sensor_freq}
\begin{tabular}{|c|c|c|}
\hline
\textbf{Aircraft Type}                                                               & \textbf{Sensor}                                                                    & \textbf{Frequency (Hz)} \\ \hline
\multirow{4}{*}{\begin{tabular}[c]{@{}c@{}}DJI Mavic Pro\\ DJI Inspire\end{tabular}} & IMU                                                                                & 200                     \\ \cline{2-3} 
                                                                                     & GPS                                                                                & 5                       \\ \cline{2-3} 
                                                                                     & Wind velocity \& direction                                                         & 5                       \\ \cline{2-3} 
                                                                                     & Battery voltage \& current                                                         & 1                       \\ \hline
\multirow{3}{*}{DJI Matrice 100}                                                     & \begin{tabular}[c]{@{}c@{}}Kinematic states\\ (IMU+GPS+Kalman Filter)\end{tabular} & 10                      \\ \cline{2-3} 
                                                                                     & Wind velocity \& direction                                                         & 10                      \\ \cline{2-3} 
                                                                                     & Battery voltage \& current                                                         & 5                       \\ \hline
\end{tabular}
\end{center}
\end{table}

Pair-wise correlations of the feature variables are illustrated in Fig.~\ref{correlation}. Most features do not have a significant linear correlation with power. The only exception is the vertical acceleration, which has a clear negative correlation with power consumption. The reason is that the generation of lift contributes a significant share of the power consumption. The change in vertical acceleration affects the required magnitude of lift force and thus influences power consumption. When the acceleration downwards increases, the required lift will decrease, and the power consumption will consequently decrease. The result of correlation analysis indicates the non-linear nature of PCM and the difficulty of accurate modeling of aircraft power consumption.

\section{Modeling and Analysis}
\label{Sec.Model}
The PCM is a regression model predicting aircraft's power as formulated in Section~\ref{Sec.Meth}. In this study, we use a stacking model to develop the PCM. The idea of stacking model is to improve model's performance by combining the strength of base-models. Therefore the performances of the base-models are crucial to the PCM. To promote accurate power prediction, we select base-models from multiple candidates based on their prediction accuracy and generalization performance, and build PCM using stacking technique.

\subsection{Machine Learning Models as Candidate Base-Models}
Candidate base-models include four popular machine learning models, namely Elastic Net regression, Random Forest regression, Gradient Boosting Regression Tree, and Multilayer Perceptron. 

\subsubsection{Elastic Net (EN) regression}
The linear regression model is one of the most commonly used regression models, which assumes a linear relationship between input and output. EN is an extension of the basic linear regression model that combines L1 and L2 regularizations to prevent overfitting and improve the model’s generalization performance. The EN regression model can be presented as:
\begin{equation}
\underset{\theta}{\operatorname{argmin}}\left\{\left(|y-X \theta|_{2}\right)^{2}+\alpha \beta|\theta|_{1}+\frac{\alpha(1-\beta)}{2}\left(|\theta|_{2}\right)^{2}\right\}
\end{equation}
where $\theta$ is the n-dimensional weight vector denoting the parameters of the model. The EN model has two hyperparameters, $\alpha > 0$ and $\beta \in [0,1]$, which manipulate the significance of L1 and L2 penalties.

\subsubsection{Random Forest (RF) regression}
RF is a typical bagging model, that combines multiple regression tree models to build a strong learner. The model builds t regression trees, each is trained by a random sample from the input with replacement, with a randomly selected subset of the features. After training, the RF model returns the average of the outputs of all t trees as the prediction. The RF model has three hyperparameters to be tuned: the number of trees t, the maximum depth of each tree, and the minimum tree leaf size. 

\subsubsection{Gradient Boosting Regression Tree (GBRT)}
Boosting is a machine learning technique that ensembles multiple weak learners (normally tree models) to become strong learners. GBRT is a popular boosting machine learning model, which shows superb performance in many contests and applications. In GBRT, the prediction is the sum of t regression trees:
\begin{equation}
\hat{y}_{i}=\sum_{k=1}^{t} f_{k}\left(x_{i}, \theta_{k}\right)
\end{equation}
where $\hat{y}_i$ is the prediction, $f_k$ is the $k^{th}$ regression tree model, and the $\theta_k$ presents the parameters of $f_k$. The training of GBRT is an iterative process, in which the $m^{th}$ step is presented as:
\begin{equation}
\hat{y}_{i}^{(m)}=\hat{y}_{i}^{(m-1)}+f_{m}\left(x_{i}\right)
\end{equation}
in which $\hat{y}_i^{(m)}$ is the sum of the output of first $m$ trees. And the $m^{th}$ tree is learning the residual of $\hat{y}_i^{(m-1)}$, by:
\begin{equation}
\underset{\theta_{m}}{\operatorname{argmin}}\left\{L\left(y_{i}, \hat{y}_{i}^{(m-1)}+f_{m}\left(x_{i}, \theta_{m}\right)\right)\right\}
\end{equation}

In this study, we chose eXtreme Gradient Boosting (XGBoost) as the implementation of GBRT. XGBoost is a very powerful open-source tool, which is scalable, portable, and allows distributed deployment. XGBoost has four hyperparameters to be tuned: the number of trees, the maximum depth of each tree, the learning rate, and the ratio of features used when constructing each tree.

\subsubsection{Multilayer Perceptron (MLP)}
MLP is a feedforward neural networks model, which has been very popular in various applications. In this study, we implement the traditional architecture of MLP: fully connected networks. We limit the number of hidden layers to three according to the results shown by Shen et al. \cite{Shen2021}. MLP has three hyperparameters to be tuned: the number of hidden layers, the neuron in each hidden layer, and the batch size in mini-batch stochastic optimization. 

\subsection{Analysis of Data Size Requirement}
The required number of samples is a key parameter balancing the model performance against the cost of data acquisition and computation. In this study, the input is tabular data, which is relatively simple compared with computer vision or natural language processing missions, and the number of features is relatively small. Hence a large dataset is not needed in obtaining a robust model. In order to determine the data size needed in the modeling of quadrotor energy consumption, a sensitivity analysis was carried out with EN, RF, and XGBoost models. To save the time cost in this analysis, the models were trained based on empirical hyperparameters. Therefore, MLP was not included in this analysis because of its high dependency on hyperparameters and the difficulty in hyperparameter tuning. 

In this analysis, we trained the three models by varying the size of the dataset. The data sizes coverage started from 100 samples and increase by 100 every time until 4500. For each data size, the original dataset was randomly sampled 50 times, and the three models were trained separately at each time. In every training, the dataset was separated into two parts, 70\% of the data forms the training set and the rest 30\% forms the testing set. We evaluated the models by their performances on the testing set by the coefficient of determination $R^2$, which is defined as:$R^{2}=1-\frac{\sum_{i}\left(y_{i}-\hat{y}_{i}\right)^{2}}{\sum_{i}\left(y_{i}-\bar{y}\right)^{2}}$, where $\bar{y}$ is the mean of $y$, and $\hat{y}_i$ is the model's prediction. 

\begin{figure}[H]
    \centering\includegraphics[width=\linewidth]{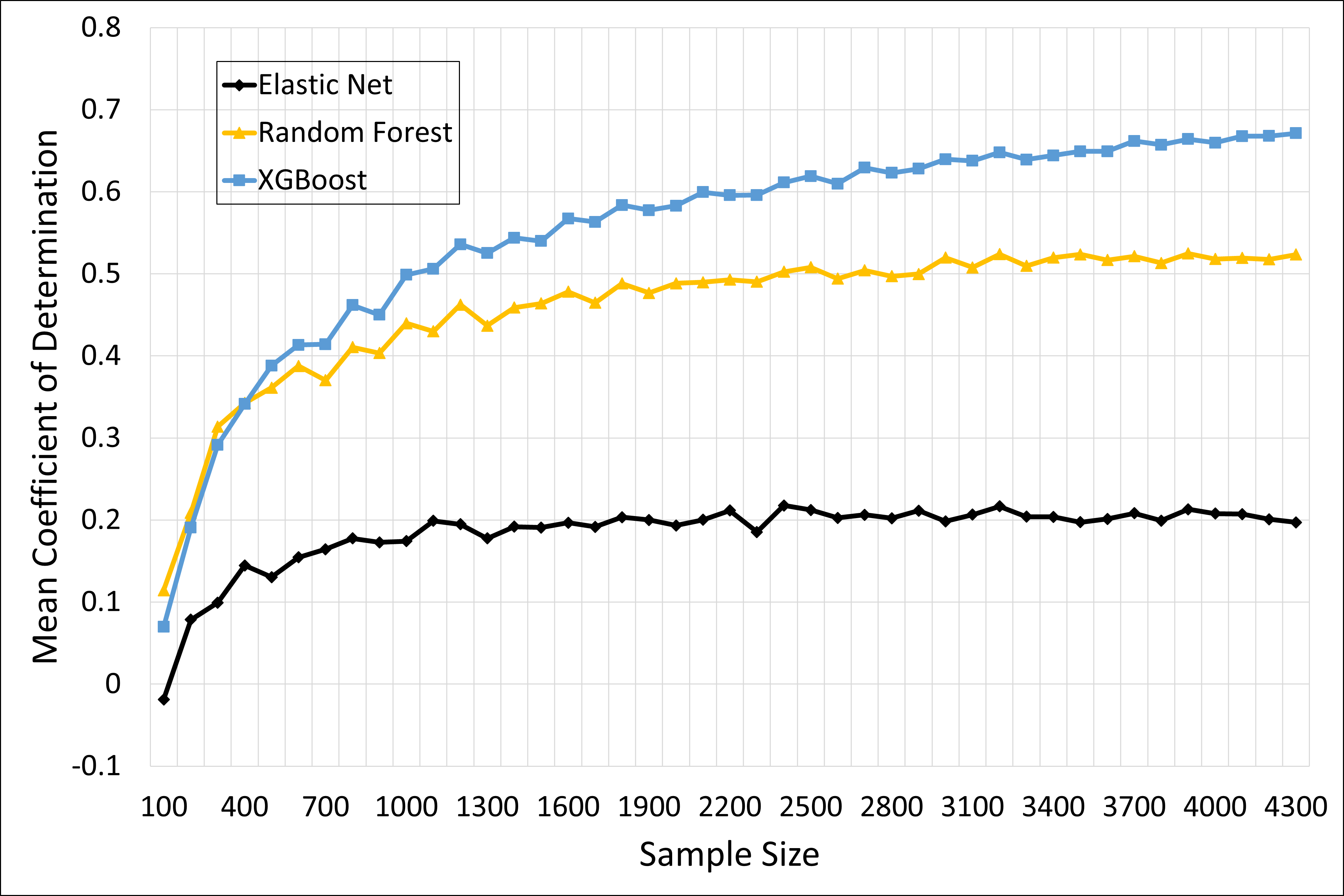}
    \caption{Mean of $R^2$ for different data size}
    \label{size_mean}
\end{figure}

\begin{figure}[H]
    \centering\includegraphics[width=\linewidth]{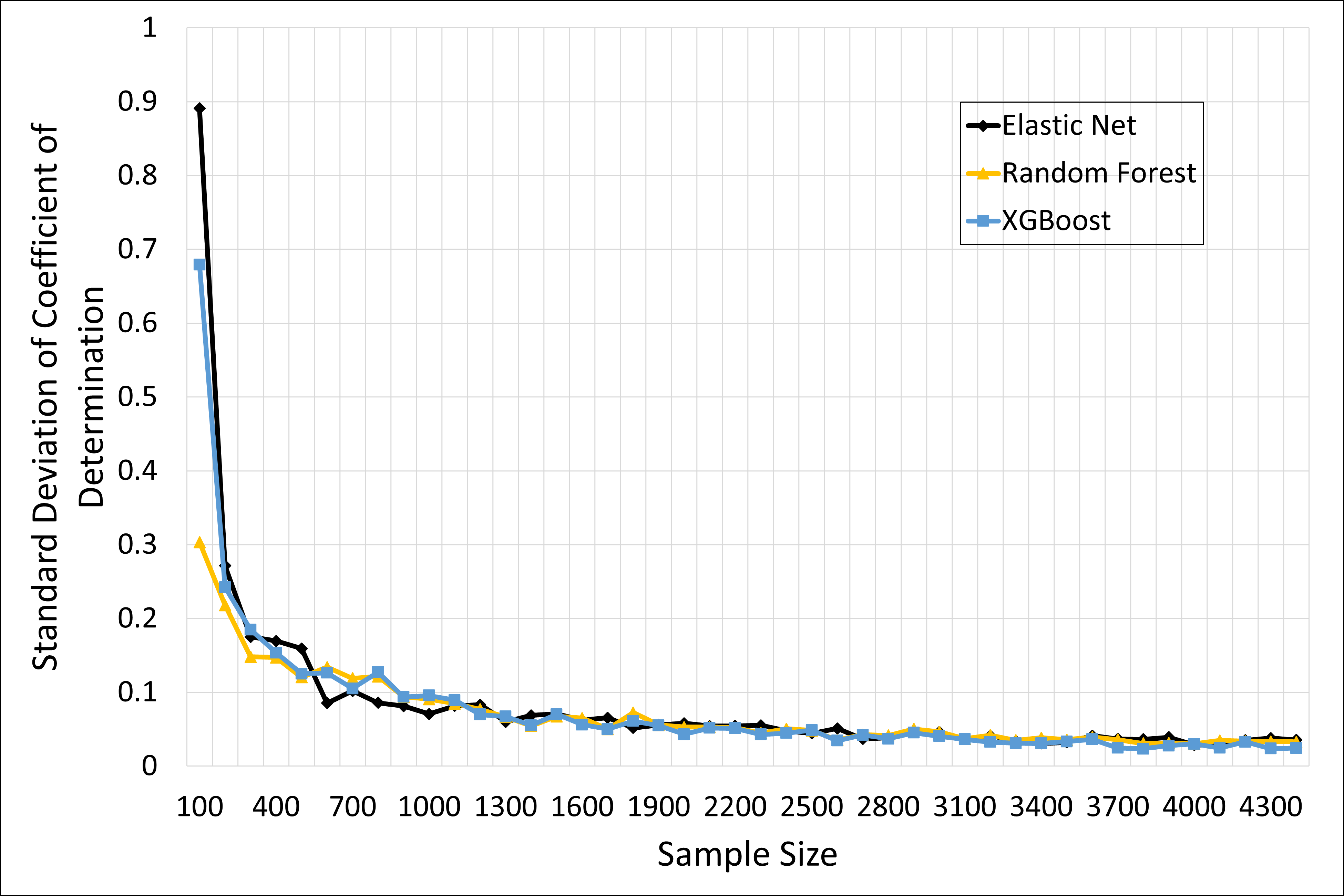}
    \caption{Standard deviation of $R^2$ for different data size}
    \label{size_std}
\end{figure}

The mean and standard deviation of $R^2$ for each data size is shown in Fig.~\ref{size_mean} and~\ref{size_std}, respectively. It can be observed that for all models the means of $R^2$  raises rapidly with the increase of data size when the data size is smaller than 1500, while the values grow slightly and tend to flat when the data size is larger than 1500. When the data size is larger than 3000, the increase in data size hardly contributes to the improvement of model performance. In terms of the standard deviation of $R^2$, the values drop significantly with the increase of data size, and fall to neglectable values when the data size is larger than 3000. This indicates that the models’ performances hardly vary with a data set larger than 3000. 

\subsection{Candidate Base-Models Tuning and Performance Analysis}
To obtain a good performance of the candidate models, hyperparameters tuning was needed. The parameters used in the tuning are shown in Table~\ref{hyperparams}. A cross-validation grid search was used to determine the best combination of the hyperparameters for each model. And models’ performances are evaluated according to the best combination of hyperparameters.

\begin{table}[H]
\begin{center}
\caption{Hyperparameters tuned for the candidate base-models.}
\label{hyperparams}
\begin{tabular}{|l|l|}
\hline
\textbf{Model}        & \textbf{Hyperparameters}                              \\ \hline
Elastic Net           & \begin{tabular}[c]{@{}l@{}}$\alpha \in [0,1]$\\ $L1Ratio \in \{ 10^{-3}, 10^{-2}, 10^{-1}\}$\end{tabular}         \\ \hline
Random Forest         & \begin{tabular}[c]{@{}l@{}}$No.Trees \in [100,1000]$\\ $MaxDepth\in[2,7]$\\ $MinSamplesLeaf \in [2^0, 2^6]$\end{tabular}     \\ \hline
XGBoost               & \begin{tabular}[c]{@{}l@{}}$No.Trees \in [100,1000]$\\ $MaxDepth\in[2,7]$\\ $LearningRate \in \{0.1, 0.3, 0.5\}$\\ $ColSampleByTree \in \{0.5, 0.8, 1\}$\end{tabular} \\ \hline
Multilayer Perceptron & \begin{tabular}[c]{@{}l@{}}$No.HiddenLayer \in \{1,2,3\}$\\ $No.NeuronInEachLayer \in [2^1, 2^8]$\\ $BatchSize \in \{2^5, 2^6, 2^7\}$\end{tabular}     \\ \hline
\end{tabular}
\end{center}
\end{table}

Besides the coefficient of determination, we also use mean squared error (MSE) and mean absolute percentage error (MAPE) to evaluate the models. The definitions of MSE and MAPE are:
\begin{equation}
    MSE =\frac{1}{n} \sum_{i=1}^{n}\left(y_{i}-\hat{y}_{i}\right)^{2}
\end{equation}
\begin{equation}
    MAPE =\frac{1}{n} \sum_{i=1}^{n}\left|\frac{\left(y_{i}-\hat{y}_{i}\right)}{y_{i}}\right|
\end{equation}

Four datasets were prepared for the modeling, including three datasets referring to the data from three different aircraft types, each having 3000 random samples, and another dataset also containing 3000 samples, in which every aircraft type has the same share. Every model was tested on each one of the four datasets separately. In every test, the dataset is split into a training set and a testing set, where the former one contains 70\% of the samples, and the latter one contains 30\%. The performances of the tuned models are shown in Table~\ref{base_model_perform}, where the best modeling performances for each dataset are presented in bold. 

\begin{table*}[htb]
\begin{center}
\caption{Performances of candidate base-models.}
\label{base_model_perform}
\resizebox{\textwidth}{!}{
\begin{tabular}{|ll|rr|rr|rr|rr|}
\hline
                                               &      & \multicolumn{2}{c|}{\textbf{Mavic Pro}}                                        & \multicolumn{2}{c|}{\textbf{Inspire}}                                          & \multicolumn{2}{c|}{\textbf{Matrice 100}}                                      & \multicolumn{2}{c|}{\textbf{Combined}}                                         \\ \cline{3-10} 
                                               &      & \multicolumn{1}{c|}{\textbf{Training}} & \multicolumn{1}{c|}{\textbf{Testing}} & \multicolumn{1}{c|}{\textbf{Training}} & \multicolumn{1}{c|}{\textbf{Testing}} & \multicolumn{1}{c|}{\textbf{Training}} & \multicolumn{1}{c|}{\textbf{Testing}} & \multicolumn{1}{c|}{\textbf{Training}} & \multicolumn{1}{c|}{\textbf{Testing}} \\ \hline
\multicolumn{1}{|l|}{\multirow{3}{*}{EN}}      & MAPE & \multicolumn{1}{r|}{0.072}             & 0.073                                 & \multicolumn{1}{r|}{0.499}             & 0.495                                 & \multicolumn{1}{r|}{0.329}             & 0.390                                 & \multicolumn{1}{r|}{0.313}             & 0.358                                 \\ \cline{2-10} 
\multicolumn{1}{|l|}{}                         & $R^2$   & \multicolumn{1}{r|}{0.170}             & 0.187                                 & \multicolumn{1}{r|}{0.019}             & 0.010                                 & \multicolumn{1}{r|}{0.313}             & 0.332                                 & \multicolumn{1}{r|}{0.878}             & 0.850                                 \\ \cline{2-10} 
\multicolumn{1}{|l|}{}                         & MSE  & \multicolumn{1}{r|}{89.336}            & 91.846                                & \multicolumn{1}{r|}{4221.966}          & 4208.584                              & \multicolumn{1}{r|}{9267.813}          & 10451.686                             & \multicolumn{1}{r|}{4670.426}          & 5643.790                              \\ \hline
\multicolumn{1}{|l|}{\multirow{3}{*}{RF}}      & MAPE & \multicolumn{1}{r|}{0.042}             & 0.050                                 & \multicolumn{1}{r|}{0.318}             & 0.362                                 & \multicolumn{1}{r|}{0.159}             & 0.176                                 & \multicolumn{1}{r|}{0.184}             & 0.252                                 \\ \cline{2-10} 
\multicolumn{1}{|l|}{}                         & $R^2$   & \multicolumn{1}{r|}{0.756}             & 0.625                                 & \multicolumn{1}{r|}{0.518}             & 0.390                                 & \multicolumn{1}{r|}{0.694}             & 0.637                                 & \multicolumn{1}{r|}{0.951}             & 0.901                                 \\ \cline{2-10} 
\multicolumn{1}{|l|}{}                         & MSE  & \multicolumn{1}{r|}{26.257}            & 42.347                                & \multicolumn{1}{r|}{2073.306}          & 2591.306                              & \multicolumn{1}{r|}{4129.109}          & 5686.904                              & \multicolumn{1}{r|}{1867.891}          & 3704.315                              \\ \hline
\multicolumn{1}{|l|}{\multirow{3}{*}{XGBoost}} & MAPE & \multicolumn{1}{r|}{\textbf{0.014}}    & \textbf{0.048}                        & \multicolumn{1}{r|}{\textbf{0.063}}    & \textbf{0.270}                        & \multicolumn{1}{r|}{\textbf{0.034}}    & \textbf{0.134}                        & \multicolumn{1}{r|}{\textbf{0.052}}    & \textbf{0.193}                        \\ \cline{2-10} 
\multicolumn{1}{|l|}{}                         & $R^2$   & \multicolumn{1}{r|}{\textbf{0.971}}    & \textbf{0.661}                        & \multicolumn{1}{r|}{\textbf{0.972}}    & \textbf{0.565}                        & \multicolumn{1}{r|}{\textbf{0.981}}    & \textbf{0.741}                        & \multicolumn{1}{r|}{\textbf{0.997}}    & \textbf{0.914}                        \\ \cline{2-10} 
\multicolumn{1}{|l|}{}                         & MSE  & \multicolumn{1}{r|}{\textbf{3.081}}    & \textbf{38.282}                       & \multicolumn{1}{r|}{\textbf{119.337}}  & \textbf{1847.281}                     & \multicolumn{1}{r|}{\textbf{254.333}}  & \textbf{4057.888}                     & \multicolumn{1}{r|}{\textbf{130.120}}  & \textbf{3240.653}                     \\ \hline
\multicolumn{1}{|l|}{\multirow{3}{*}{MLP}}     & MAPE & \multicolumn{1}{r|}{0.058}             & 0.063                                 & \multicolumn{1}{r|}{0.472}             & 0.473                                 & \multicolumn{1}{r|}{0.300}             & 0.354                                 & \multicolumn{1}{r|}{0.286}             & 0.327                                 \\ \cline{2-10} 
\multicolumn{1}{|l|}{}                         & $R^2$   & \multicolumn{1}{r|}{0.495}             & 0.406                                 & \multicolumn{1}{r|}{0.146}             & 0.117                                 & \multicolumn{1}{r|}{0.427}             & 0.420                                 & \multicolumn{1}{r|}{0.889}             & 0.864                                 \\ \cline{2-10} 
\multicolumn{1}{|l|}{}                         & MSE  & \multicolumn{1}{r|}{54.436}            & 67.136                                & \multicolumn{1}{r|}{3673.654}          & 3752.189                              & \multicolumn{1}{r|}{7728.984}          & 9078.724                              & \multicolumn{1}{r|}{4254.765}          & 5110.980                              \\ \hline
\end{tabular}}
\end{center}
\end{table*}
Amongst the four models, EN has the lowest performance in all MSE, $R^2$ and MAPE, indicating a highly non-linear relationship between selected features and the aircraft’s power of energy consumption. EN is not capable of modeling the relationship and the result turns to be strongly underfitting. XGBoost outperforms the other three models on all four datasets. On the training sets, XGBoost has an $R^2$ of more than 0.95, and an MAPE of less than 0.05, showing that the model fits the training data very well. However, there is a significant drop in the values of the performance indicators from the training set to the testing set. Although the test was carried out by increasing the complexity of XGBoost, and the result indicates that the model is not overfitting, there is a clear fact that XGBoost lacks generalization performance. The second-best model is RF. Compared with XGBoost, RF shows a better generalization performance, as its performance shrinks slighter than XGBoost. Comparing the four datasets, it can be found that all models have small MSE and MAPE values on the dataset of Mavic Pro. This might be because of the small value of the power of the aircraft, as it is the lightest aircraft among the three and has the smallest required energy in changing its kinetic state.

\subsection{Development of the Stacking Model}

Based on the modeling results, XGBoost is selected as one of the base-models because of its supreme performance. And RF was also included to ameliorate the stacking model’s generalization performance. The meta-model used in this study is a linear regressor. Five-fold cross-validation was applied to train the base-models, namely RF and XGBoost, and the predictions of base-models were used to train the meta-model. Finally the entire stacking model will be validated on the testing set. 

\section{Modeling Result Analysis}
\label{Sec.Result}
\begin{figure}[H]
    \centering\includegraphics[width=\linewidth]{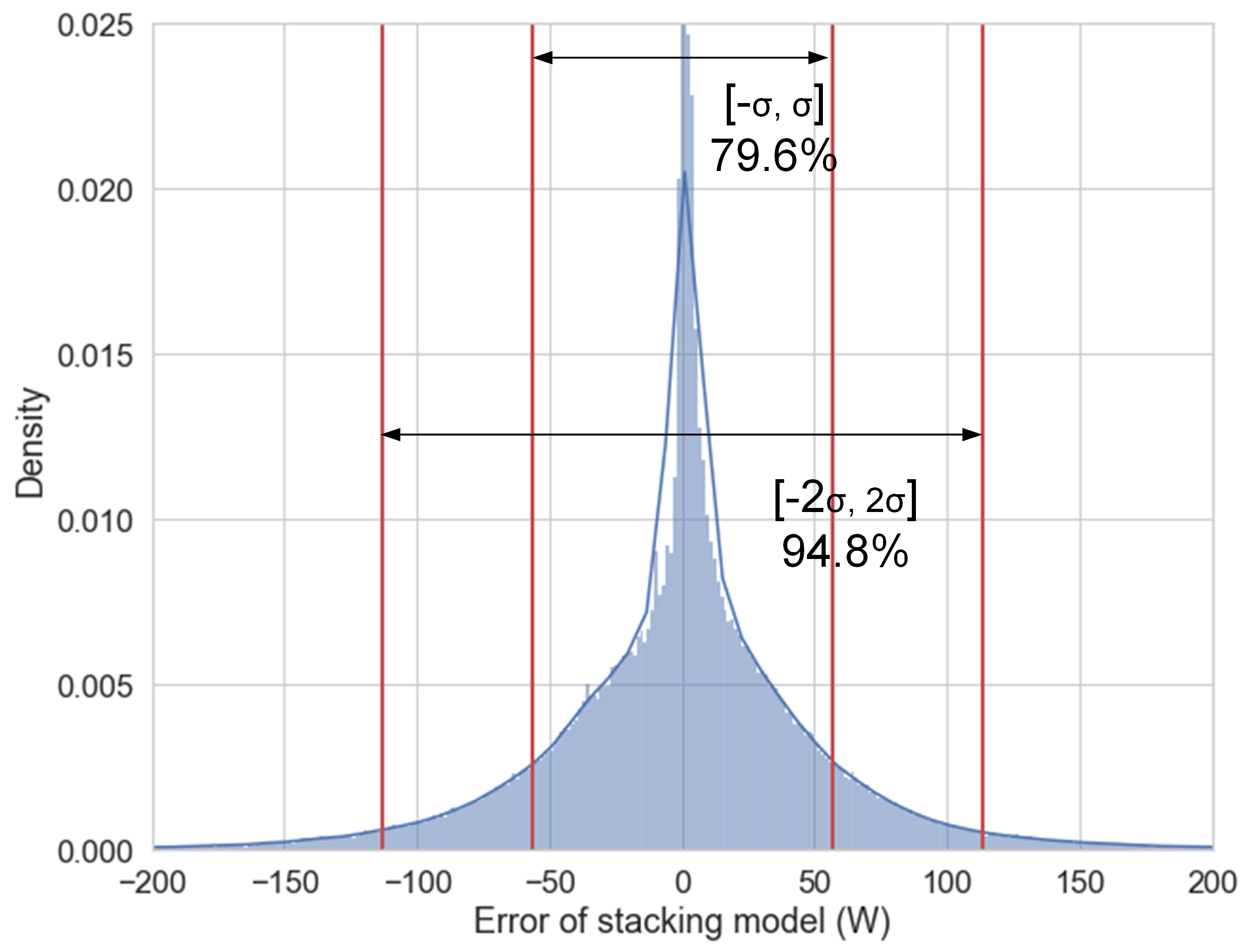}
    \caption{Density distribution of prediction errors of the stacking model}
    \label{hist_error}
\end{figure}

To obtain a clear understanding of the prediction uncertainty of our model, the density distribution of error is illustrated in Fig.~\ref{hist_error}. A symmetric single peak distribution is observed, where the mean is zero and the standard deviation ($\sigma$) is 56.7 W. While the long tail effect makes the distribution non-normal, the distribution has a good statistical property. Almost 80\% of the errors fall within the interval of $\pm \sigma$, and approximately 95\% are within $\pm 2\sigma$.

\begin{figure}[H]
    \centering\includegraphics[width=\linewidth]{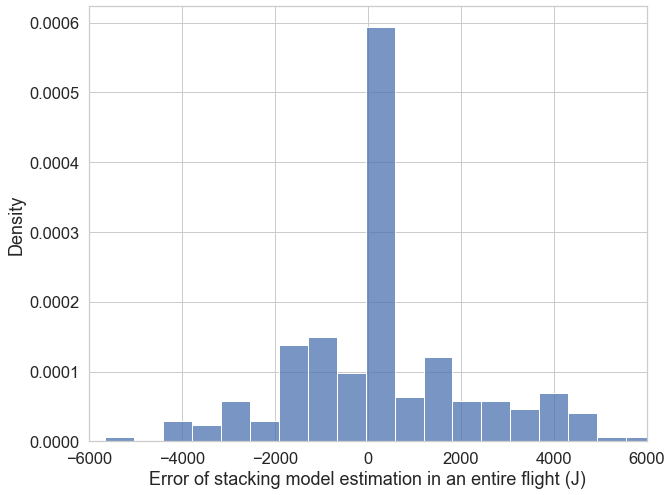}
    \caption{Density distribution of prediction error accumulated in an entire flight}
    \label{accum_error}
\end{figure}

The symmetrical distribution of error leads to a superior performance in the prediction of power consumption for an entire flight mission. The probability density distribution of prediction error for a flight is illustrated in Fig.~\ref{accum_error}. In 82.8\% of the flights, the prediction error is  within the interval of $\pm 2000$ J. Given the fact that the capacity of standard battery for Matrice 100 is 467,856 J (129.96 Wh), a prediction error of less than 0.5\% of battery's capacity is totally acceptable.

\subsection{Comparing with the State-of-the-Art}
Two models from the literature are used for comparison. The first one is the model of Hong et al.~\cite{Hong2020}, who used a fully connected MLP with 35 hidden layers, each layer has 44 neurons. The second one is the model of Choudhry et al.~\cite{Choudhry2021}, who used a TCN with residual connection. The dataset of Matrice 100 was used for the comparison because of two reasons. Firstly, as reported by Hong et al.~\cite{Hong2020}, the onboard sensor of DJI aircraft is not very accurate and may lower the reliability of the modeling outcome. In the acquisition of Matrice 100 dataset, the researchers used self-built sensing system to obtain a better accuracy~\cite{Rodrigues2021}. Secondly, both MLP and TCN requires a large training dataset to guarantee their convergences, and only Matrice 100 dataset is sufficient. The Matrice 100 dataset contains 279 flights. We use 80\% of the data, i.e. the data of 233 flights to train the models. And data of the left 56 flights were used to validate the model.

\begin{figure*}[htb]
      \centering
        \subfloat[Power prediction for a flight mission by stacking model compared with ground truth]{\includegraphics[width=\textwidth]{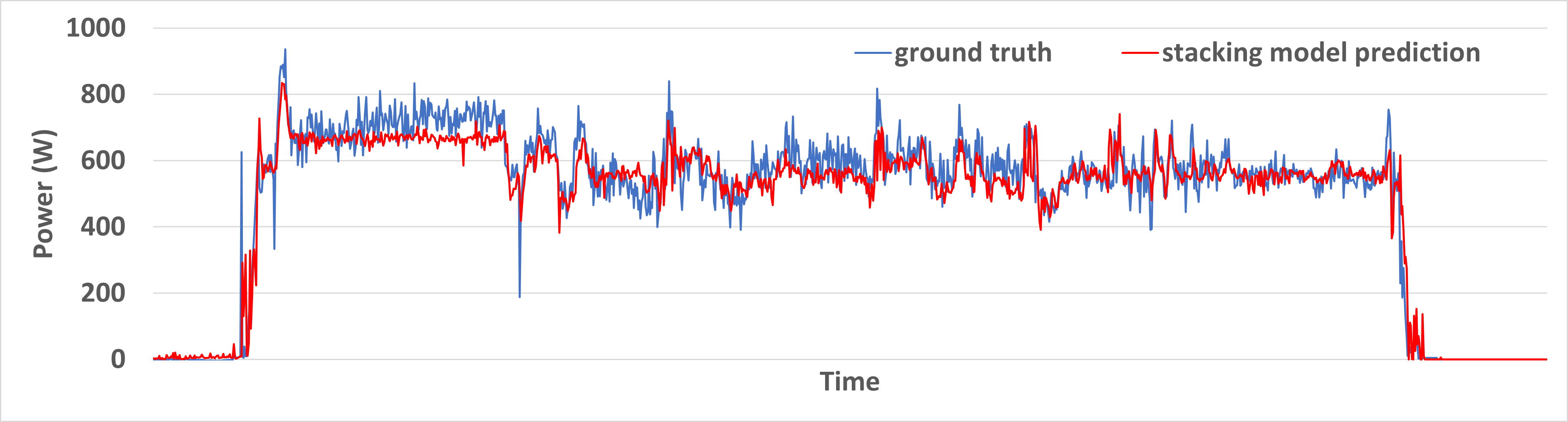}} \\
    	\subfloat[Power prediction for a flight mission by MLP compared with ground truth]{\includegraphics[width=\textwidth]{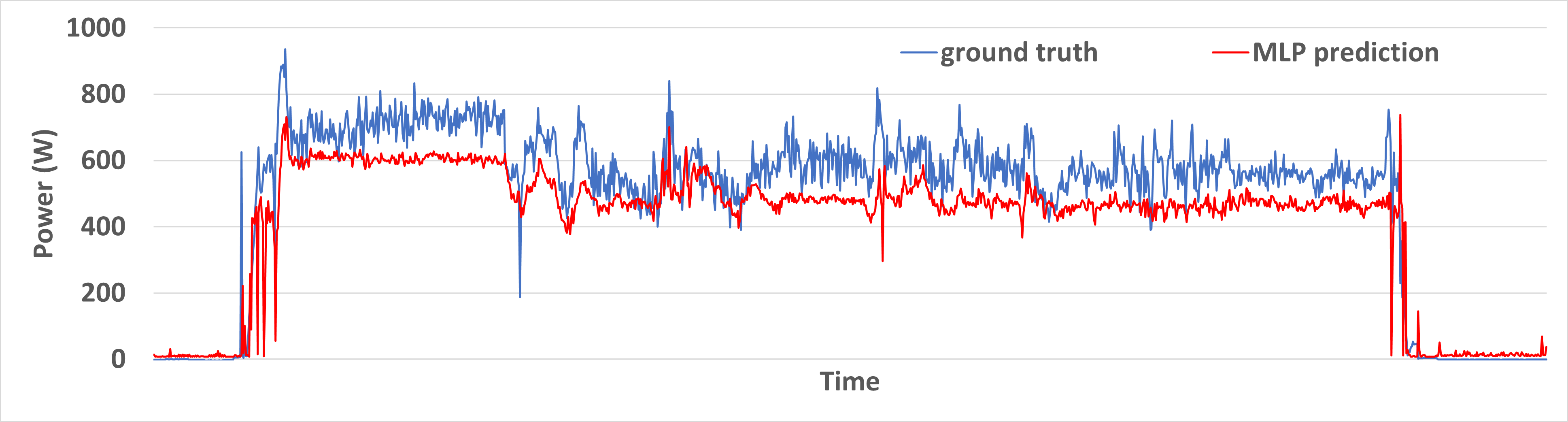}} \\
        \subfloat[Power prediction for a flight mission by TCN compared with ground truth]{\includegraphics[width=\textwidth]{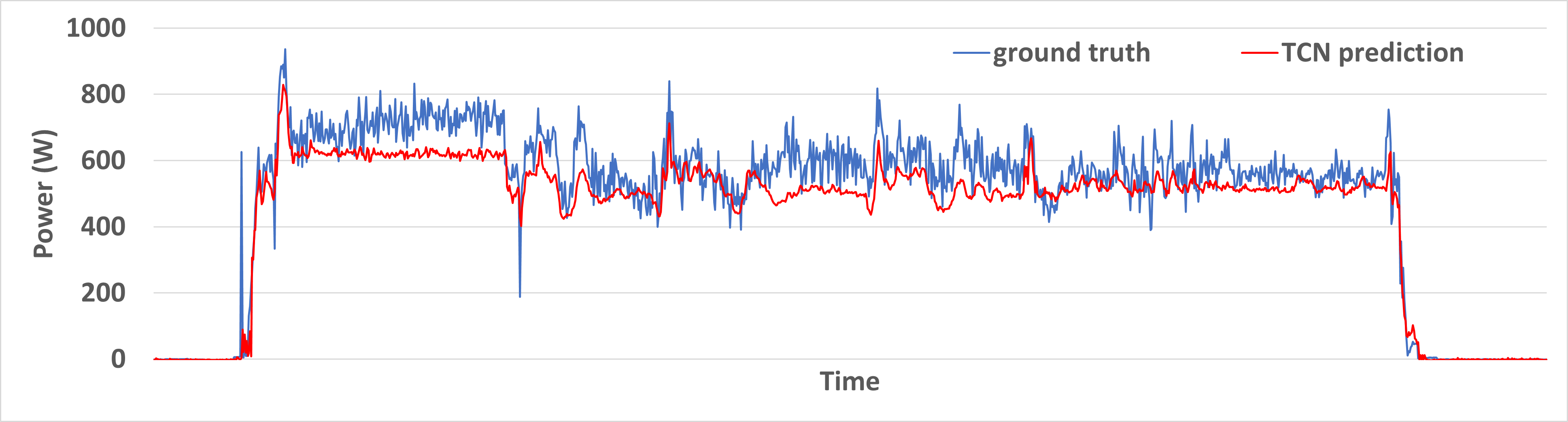}} \\
      \caption{Comparison between ground truth and models’ predictions in a flight randomly picked from the testing data}
    	\label{whole_comp}
    	\vspace{0.2in}
\end{figure*}

To obtain a clear vision of the accuracy performance of the models in predicting the power during a flight mission, a comparison was made to visualize the difference between prediction and ground truth. The result is illustrated in Fig.~\ref{whole_comp}, where Fig.~\ref{whole_comp}(a), (b), and (c) compare the ground truth with the prediction of our stacking model, MLP, and TCN, respectively. Amongst the three models, our model has the best prediction of the ground truth, especially in the approximation of the higher frequent variations of the power.

\begin{figure}[H]
    \centering\includegraphics[width=0.7\linewidth]{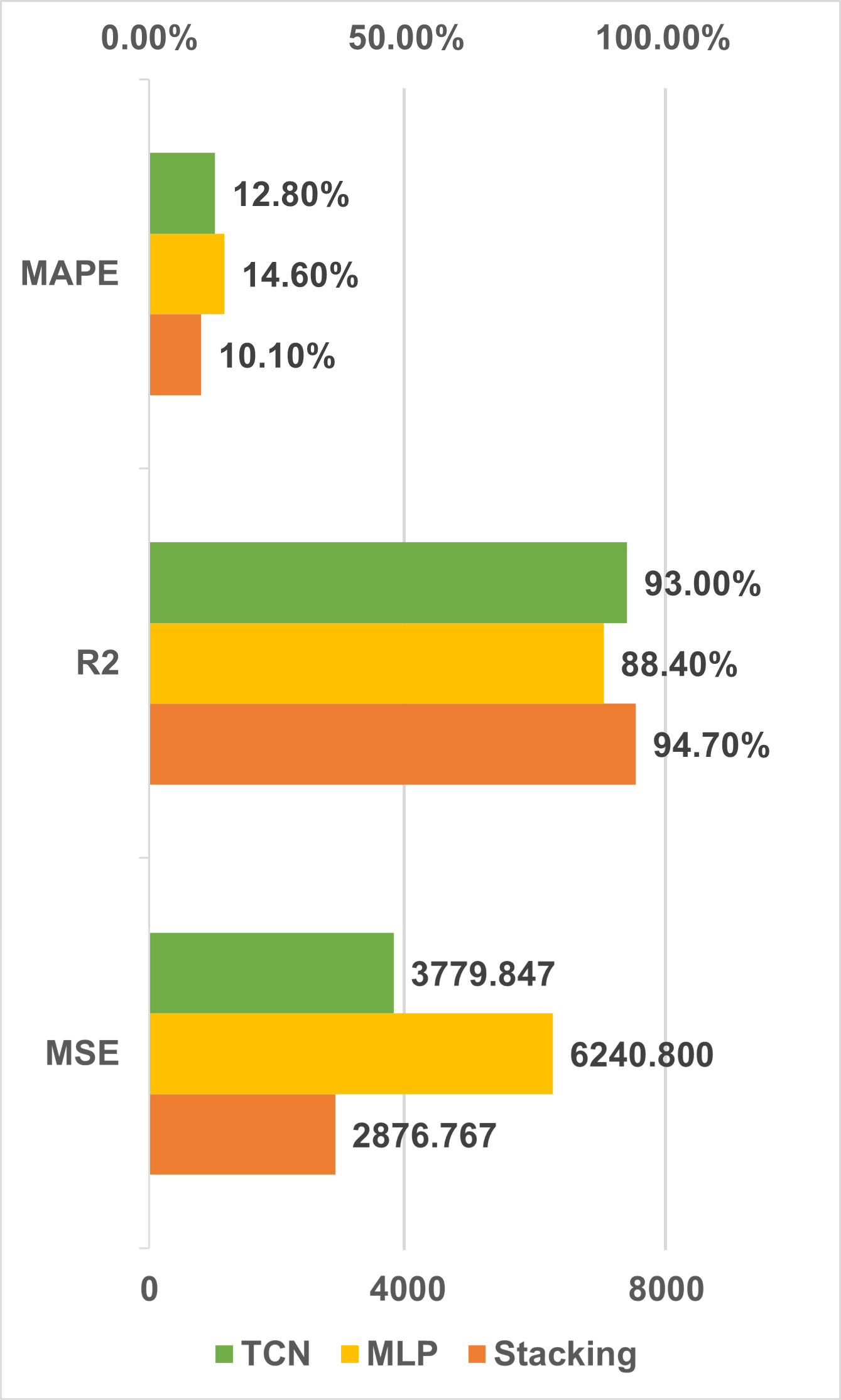}
    \caption{Statistical results of the stacking model on test dataset in comparison with MLP and TCN}
    \label{comp}
\end{figure}

Statistical results of the comparison between our model and the other two models are shown in Fig~\ref{comp}. Our model has the highest $R^2$ value, and the lowest MSE and MAPE errors. In terms all of these three metrics, our model has the best performance comparing with existing models. 

\section{Conclusions}
\label{Sec.Con}
In this study, a framework for estimating the power of energy consumption by a quadrotor was established using the ensemble learning method. Three datasets were employed in this work, including flight test recordings of two aircraft and an open-source dataset. Four machine learning models, namely EN, RF, XGBoost, and MLP, were tuned and tested with the datasets. And a stacking model was designed, using RF and XGBoost, the two models showing better performance, as the base-models, and a linear regressor as the meta-model. Finally, the performance of the stacking model was analyzed by both power curve approximation performance and statistical performance metrics. Results show that the ensemble learning model developed in this study has a good performance in estimating the power of energy consumption of a quad-rotorcraft. Error analysis shows that roughly 80\% of instantaneous power estimation errors are within one standard deviation, and less than 0.5\% error in the power prediction for an entire flight can be expected with a confidence of more than 80\%. Compared with the state-of-the-art models, our model has the best performances in three metrics including MAPE, $R^2$, and MSE. The model provides a robust tool for flight operators in the planning and management of flight missions. 

A basic principle in this study is to guarantee the model’s scalability. Compared with popular deep learning models that require a large set of training data, training the stacking model developed in this study needs only several thousand samples. Due to the quick evolution and the system simplicity in the eVTOL aircraft types, the industry will face the challenge of significant aircraft diversity. The characteristic of requiring a small training dataset will reduce the cost of time and resource in the implementation of new aircraft types, and greatly improve the flexibility of flight management. Introducing more features, especially the detailed specifications of aircraft, to the model might also obtain a better performance of prediction. But it will also increase the difficulty of applying the model in scale.

\bibliographystyle{IEEEtran}
\bibliography{PCM.bib}

\end{document}